\documentclass[11pt]{article}

\usepackage[preprint]{acl}

\usepackage{times}
\usepackage{latexsym}

\usepackage[T1]{fontenc}
\usepackage[utf8]{inputenc}

\usepackage{microtype}

\usepackage{inconsolata}

\usepackage{graphicx}
\usepackage{algorithmic}
\usepackage[linesnumbered,ruled,vlined]{algorithm2e}
\usepackage{amsmath}
\usepackage{multirow}
\usepackage{booktabs}
\usepackage[most]{tcolorbox}
\usepackage{enumitem}   
\usepackage{circledsteps}
\usepackage{xcolor}
\usepackage{tikz}
\usepackage{amssymb}

\newcommand{\circnum}[1]{%
  \tikz[baseline=(C.base)]{
    \node[
      fill=black,
      circle,
      inner sep=1.2pt
    ] (C) {\textcolor{white}{\footnotesize\textbf{#1}}};
  }\,
}

\title{DeepEra: A Deep Evidence Reranking Agent for Scientific  Retrieval-Augmented Generated Question Answering}

\author{
 \textbf{Haotian Chen\textsuperscript{1}},
  \textbf{Qingqing Long\textsuperscript{1}\thanks{Corresponding Authors}},
 \textbf{Siyu Pu\textsuperscript{1}},
 \textbf{Xiao Luo\textsuperscript{2}},
 \textbf{Wei Ju\textsuperscript{3}},\\
 \textbf{Meng Xiao\textsuperscript{1}},
 \textbf{Yuanchun Zhou\textsuperscript{1}},
 \textbf{Jianghua Zhao\textsuperscript{1}},
 \textbf{Xuezhi Wang\textsuperscript{1}$^*$}
 \\
 \textsuperscript{1}Computer Network Information Center, Chinese Academy of Sciences \\
  \textsuperscript{2}University of Wisconsin–Madison,
  \textsuperscript{3}Peking University
}

\begin{document}
\maketitle
\begin{abstract}
With the rapid growth of scientific literature, scientific question answering (SciQA) has become increasingly critical for exploring and utilizing scientific knowledge. Retrieval-Augmented Generation (RAG) enhances LLMs by incorporating knowledge from external sources, thereby providing credible evidence for scientific question answering. But existing retrieval and reranking methods remain vulnerable to passages that are semantically similar but logically irrelevant, often reducing factual reliability and amplifying hallucinations.
To address this challenge, we propose a \textbf{Deep} \textbf{E}vidence \textbf{R}eranking \textbf{A}gent (\textbf{DeepEra}) that integrates step-by-step reasoning, enabling more precise evaluation of candidate passages beyond surface-level semantics. To support systematic evaluation, we construct \textit{SciRAG-SSLI (Scientific RAG–Semantically Similar but Logically Irrelevant)}, a large-scale dataset comprising about 300K SciQA instances across 10 subjects, constructed from 10M scientific corpus. The dataset combines naturally retrieved contexts with systematically generated distractors to test logical robustness and factual grounding. Comprehensive evaluations confirm that our approach achieves superior retrieval performance compared to leading rerankers. 
To our knowledge, this work is the first to comprehensively study and empirically validate innegligible \textit{SSLI} issues in two-stage RAG frameworks.
\end{abstract}

\section{Introduction}
Scientific question answering plays a critical role in advancing scientific discovery~\cite{kraemer2025artificial,wang2025integrating}.
For instance, molecular biologists studying CRISPR gene editing must review prior work to assess off target effects~\cite{hsu2013dna}, while during global health emergencies such as the COVID-19 pandemic, clinicians must rapidly synthesize large volumes of newly published studies to support evidence based decisions~\cite{zammarchi2024scientometric}.
Recent large language models (LLMs) exhibit strong reasoning and language understanding abilities, but remain prone to hallucinations, outdated knowledge, and insufficient evidential grounding~\cite{farquhar2024detecting,guo2025deepseek,augenstein2024factuality}. 
Retrieval-Augmented Generation (RAG) mitigates these issues by grounding generation in up-to-date external scientific literature, improving accuracy and reliability, especially in scientific domains where evidence fidelity is critical~\cite{wu2025medical,chen2025scirerankbench,yang2024rag}.

\begin{figure}[t]
    \centering
    \includegraphics[width=0.9\columnwidth]{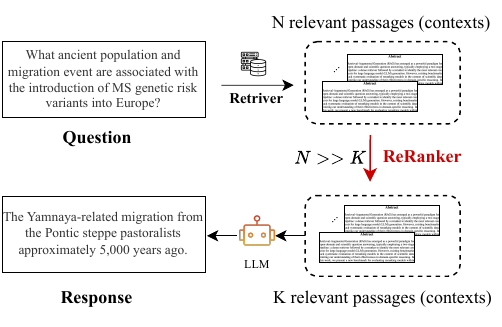}
    \vspace{-3mm}
    \caption{
    The \textit{two-stage} RAG-LLM pipeline. 
    }
    \label{fig:rag}
    \vspace{-6mm}
\end{figure}

Despite its effectiveness in grounding language model outputs, RAG still suffers from noisy and weakly relevant retrieved contexts, which can degrade answer quality~\cite{yang2024crag,augenstein2024factuality}. To address these challenges, rerank models are introduced to rerank retrieved passages by relevance. The two-stage pipline is shown in Figure ~\ref{fig:rag}. Dense cross-encoder rerankers~\cite{zhuang2025rank,dong2024don}, such as BGE~\cite{chen2024bge}, BCE~\cite{youdao_bcembedding_2023}, and Jina~\cite{jina}, score query document pairs to promote the most informative passages, thereby improving answer fidelity and robustness in RAG systems.

\begin{figure}[t]
    \centering
    \includegraphics[width=\columnwidth]{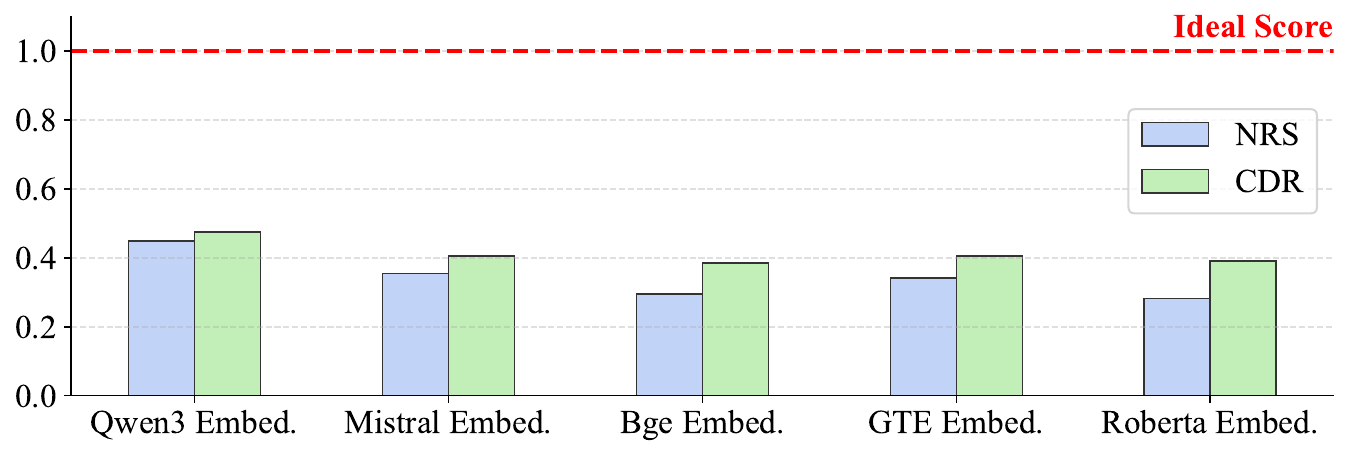}
    \vspace{-6mm}
    \caption{Retrieval performance of embedding models with similar but logically irrelevant passages.}
    \vspace{-6mm}
    \label{fig:into_embedding}
\end{figure}

\paragraph{Challenges and Empirical Validation.} 
\textbf{ (1) Theoretical Upper Bound.}
Current RAG methods, including both embedding-based retrieval and reranking, primarily rely on vector similarity. Such similarity can be misleading, as passages may appear close in embedding space due to lexical overlap while remaining logically irrelevant to the query.~\cite{guo-liang-2025-transform} These semantically similar yet uninformative passages introduce noise and undermine the reliability of evidence provided to LLMs~\cite{augenstein2024factuality,fang2024enhancing}, limiting the robustness of existing RAG systems. 
\textbf{(2) Empirical Validation.}
To empirically examine this limitation, we generate semantically similar but logically irrelevant passages using LLMs and randomly inject them into the retrieval database. We evaluate several representative embedding models, including Qwen3-Embedding-8B~\cite{zhang2025qwen3}, E5-Mistral-7B-Instruct~\cite{choi2024linq}, BGE-M3~\cite{chen2024bge}, GTE-Large-v1.5~\cite{li2023towards}, and all-roberta-large-v1~\cite{zuo2025enhancement}. We introduce two metrics: Noise Robustness Score (NRS), measuring the proportion of non-distractor passages retrieved, and Context Discrimination Rate (CDR), capturing the ability to distinguish original passages from their distractors. As shown in Fig.~\ref{fig:into_embedding}, \textbf{both metrics fall substantially below the ideal}, indicating that current \textbf{embedding models struggle to filter semantically similar but logically irrelevant contexts}. These results expose a fundamental limitation of embedding based RAG and motivate rerankers that model logical relevance beyond surface similarity.

\paragraph{Proposed Method.} To bridge the identified gap, we propose \textbf{Deep} \textbf{E}vidence \textbf{R}eranking \textbf{A}gent (\textbf{DeepEra}), an agentic reranker that leverages the reasoning capabilities of LLMs to assess candidate passages beyond surface-level similarity. Unlike conventional embedding-based rerankers, DeepEra evaluates contexts based on their semantic relevance, logical consistency, and evidential grounding with respect to the query, mitigating the impact of semantically similar but misleading passages and improving the robustness and reliability of scientific RAG.
Due to the lack of benchmarks containing semantically similar but logically irrelevant negatives, we further construct a dataset, SciRAG-SSLI, by mixing naturally retrieved scientific passages with LLM-generated distractors to ensure both realism and controlled difficulty.

We summarize our contributions as follows:
\circnum{1}
To our knowledge, this work is \textbf{the first} to 
comprehensively and empirically validate innegligible \textbf{\textit{SSLI} issues in two-stage RAG} frameworks.
\circnum{2}
We construct a new dataset, \textbf{SciRAG-SSLI}, which combines naturally retrieved passages with LLM-generated distractors to provide controlled yet realistic evaluation scenarios for scientific QA.
\circnum{3} 
We propose DeepEra for scientific retrieval-augmented generation, designed to dynamically reason over candidate passages and filter misleading contexts.
\circnum{4} Extensive experiments show that DeepEra performs better than other rerankers, achieving up to \textbf{8\% relative improvements} in \textbf{retrieval robustness and answer accuracy}.

\begin{figure*}[t]
    \centering
    \includegraphics[width=\textwidth]{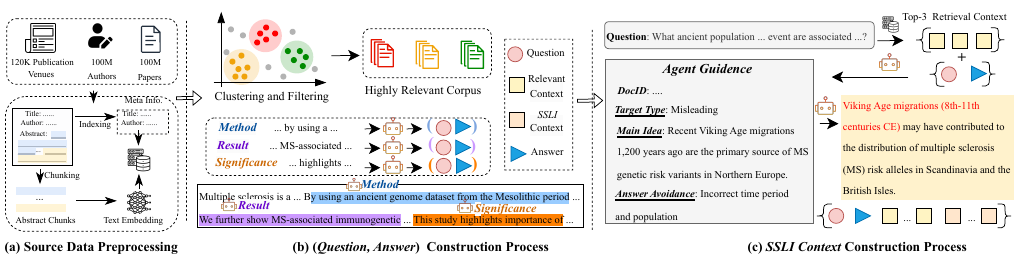}
    \vspace{-9mm}
    \caption{
Dataset construction pipeline.
Scientific abstracts and metadata are collected and clustered to retain contextually related documents.
Structured information (\textit{Method}, \textit{Result}, \textit{Significance}) is extracted to generate QA pairs.
For each question, \textit{SSLI} contexts are further generated via LLM-guided instructions to create semantically similar but logically irrelevant distractors.
    }
    \vspace{-4mm}
    \label{fig:dataset}
\end{figure*}

\section{Related Work}

Our work relates to three research lines: scientific question answering, retrieval augmented generation with reranking, and agentic scientific QA.
Although diverse rerankers ranging from dense and sparse to listwise and LLM-based methods have been proposed~\cite{youdao_bcembedding_2023,chen2024bge,jina}, most rely on surface-level semantic similarity and remain vulnerable to semantically similar but logically irrelevant passages.
Meanwhile, agent based scientific QA enables multi step retrieval and reasoning but typically lacks explicit reranking to filter misleading evidence.~\cite{lala2023paperqa,pasa2025}
Our Deep Evidence Reranking Agent bridges these directions by introducing agentic reasoning into reranking, explicitly modeling logical relevance and evidential support.
A detailed review of related work is provided in \textbf{Appendix~\ref{app:related_work}}.

\section{Methodology}
\label{sec:Methodology}
\noindent\textbf{Problem Definition}. In this paper, we consider the task of scientific question answering within the RAG framework, which consists of three main stages. 
Given a scientific query $q$, the goal is to generate an answer $a$ that is supported by credible and contextually relevant evidence. 
First, a retriever searches over a large scale corpus $\mathcal{C}$ and returns a candidate set of passages (evidence)
$\mathcal{P} = \{p_1, p_2, \ldots, p_n\}, \ \mathcal{P} \subset \mathcal{C}$, 
based on their vector similarity to the query. 
Next, to improve the quality of the generator’s answer, a reranker $R(q,p)$ is applied to rescore the candidate evidence set, and the top-$k$ ($k \ll N$) evidence with the highest scores are retained as the refined context set 
$\mathcal{P}^* = \text{Top-}k(\{R(q,p_i)\}_{i=1}^n)$. 
Finally, an LLM takes the query $q$ together with the refined set $\mathcal{P}^*$ and produces the final answer as 
$a = \text{LLM}(q,\mathcal{P}^*)$. 

\begin{figure*}[htbp]
    \centering
    \includegraphics[width=\textwidth]{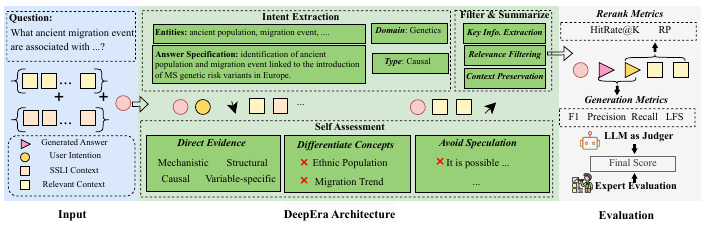}
    \vspace{-10mm}
    \caption{
    Overview of DeepEra. 
    DeepEra first performs \textbf{\textit{Intention Recognition}} to extract structured query representations. 
    Retrieved candidate passages are then scored in the \textbf{\textit{Relevance Assessment}} stage using an LLM-based function to prioritize mechanistically or causally relevant evidence. 
    Finally, passages exceeding a relevance threshold are \textbf{\textit{filtered and summarized}} to produce a compact evidence set for downstream answer generation.
    }
    \vspace{-3mm}
    \label{fig:rerank-agent}
\end{figure*}

\subsection{Dataset Construction}

To effectively evaluate our proposed framework, we constructed two datasets tailored for scientific question answering and reranking tasks, as shown in Fig.~\ref{fig:dataset}. Both datasets are built by first generating question-answer (QA) pairs from scientific documents and then defining two settings to control the retrieval difficulty. 

\paragraph{QA Generation.} We adopted two approaches for QA generation. 
The first method follows the \textit{LMQG} framework ~\cite{lmqg}, which employs LLMs to automatically generate QA pairs from scientific papers. 
The second approach uses an LLM with carefully designed prompts to generate QA pairs, for ensuring reproducibility and domain customization. 
As illustrated in Fig.~\ref{fig:dataset}, we collect scientific papers from open-access repositories~\cite{priem2022openalex} and segment abstracts into sentence-preserving chunks. These chunks are embedded and stored with associated metadata in a structured database. We then filter weakly connected documents via clustering and extract structured information (e.g., \textit{methods}, \textit{results}, and \textit{significance}) using LLMs. Based on this structured information, QA pairs are generated to align with scientific QA tasks. Together, these approaches produce two datasets for downstream experiments.

\paragraph{Base and \textit{SSLI} Settings.} For each dataset, we define two kinds of settings: 
(1) \textbf{Base}, where questions are paired with contexts directly retrieved from the scientific database, and 
(2) \textbf{\textit{SSLI}} (\textit{Semantically Similar but Logically Irrelevant}), where additional distractor passages (context) are injected into the candidate passages (context). 
For each QA pair, we first retrieve the top-3 contexts $C$ by embedding similarity. The query $Q$, golden answer $A$, and retrieved contexts $C$ are then provided to an LLM, which outputs structured guidance to control distractor generation. This guidance specifies the intended distractor type (e.g., misleading, background, irrelevant), thematic focus, and instructions to avoid including the golden answer. Guided by this, the LLM generates distractors that are semantically coherent with the query and retrieved passages but logically inconsistent with the ground truth answers.

\subsection{Model: DeepEra}

To enhance the quality and reliability of evidence provided to downstream generation, we propose the Deep Evidence Reranking Agent (\textbf{DeepEra}), which systematically integrates query understanding, structured scoring, and evidence summarization. 
As shown in Figure ~\ref{fig:rerank-agent}. DeepEra is designed to operate in three sequential stages, explicitly modeling both the semantic and logical relevance of candidate passages with respect to the scientific query. 

\paragraph{Intention Recognition} 
Given a scientific question $q$, we first employ an LLM to perform intention recognition. The model extracts a structured representation of the query, denoted as $I(q)=\{\textit{topic}, \textit{entity\_type}, \textit{intent}, \textit{expected\_answer\_type}\}$. 
Here, \textit{topic} identifies the overarching domain (e.g., molecular biology, epidemiology), \textit{entity\_type} specifies the central scientific entities (e.g., ``cell type: lymphocytes'' or ``virus strain: SARS-CoV-2''), \textit{intent} denotes the cognitive or reasoning task (definition, mechanism, comparison, causal inference, or functional role), and \textit{expected\_answer\_type} indicates the preferred form of the answer (e.g., numeric, categorical, procedural, or descriptive). By structuring the query in this manner, DeepEra ensures that downstream reranking focuses on passages that are scientifically relevant and contextually aligned, rather than superficially matching keywords or common terminology.

\paragraph{Relevance Assessment} 
Each retrieved passage $p_i \in \mathcal{P}$ is evaluated against the structured query representation using an LLM-based scoring function. The scoring prompt enforces rigorous scientific selection criteria, ensuring that only passages offering direct mechanistic, structural, or causal evidence are prioritized. Specifically, passages must explicitly address variable-specific relationships or provide data-driven results to achieve a high score, while speculative or tangentially related content is filtered out. Formally, the relevance of a passage is defined as $s(q, p_i) = \text{Score}(q_{\textit{struct}}, p_i)$, where $q_{\textit{struct}}$ denotes the structured representation of the query. The resulting \textit{RelevanceScore} lies within $[0,1]$, with high scores reserved for passages containing critical experimental findings or structural derivations, whereas redundant, vague, or conceptually mismatched passages are assigned low scores. This strict differentiation of concepts, models, variables, and domains prevents superficial lexical matches from being misclassified as relevant, aligning the scoring process with reasoning.

\paragraph{Evidence Filtering and Summarization}
Passages are filtered strictly according to their \textit{RelevanceScore}, and only those exceeding a predefined threshold $\tau$ are retained, i.e., $\mathcal{P}^* = \{p_i \mid s(q, p_i) \geq \tau \}$. This ensures that semantically similar but logically irrelevant passages are excluded, leaving only trustworthy evidence aligned with the structured query. To further reduce redundancy and control the input length to the generator LLM, each retained passage is summarized into a concise form conditioned on the query representation. The summarization process follows three key principles: (1) \textbf{Key Information Extraction}: identify only the content directly addressing the scientific question; (2) \textbf{Context Preservation}: preserve original terminology and entities without paraphrasing, thereby avoiding semantic drift; and (3) \textbf{Conciseness}: produce a logically clear summary limited to one or two sentences. 
This produces a high-precision evidence set that maximizes relevance, reduces noise from misleading or tangential passages, and facilitates robust answer generation. The pseudocode of DeepEra is shown in Algo. ~\ref{alg:agentic_reranker}.

\begin{algorithm}[t]
\small
\caption{DeepEra (Deep Evidence Reranking Agent)}
\label{alg:agentic_reranker}
\KwIn{Question $q$; retrieved passages (candidate evidence set) $\mathcal{P}=\{p_1,\dots,p_n\}$; threshold $\tau$}
\KwOut{Final evidence list $\mathcal{E}$}

\BlankLine
\textbf{Step 1: Intention Recognition} \\
$I(q) \gets \text{LLM}_{\text{intent}}(q)$ \\
$I(q) = \{\;t, e, i, a\;\}$ \tcp*{$t$: topic, $e$: entity\_type, $i$: intent, $a$: expected\_answer\_type}

\BlankLine
\textbf{Step 2: Relevance Assessment} \\
\For{$p_i \in \mathcal{P}$}{
    $s(q,p_i) \gets \text{LLM}_{\text{score}}(I(q), p_i)$ \\
    \tcp*{Scoring by structured match: high if $p_i$ contains direct evidence}
}
$\mathcal{S} = \{ (p_i, s(q,p_i)) \mid i=1,\dots,n \}$ \\
$\mathcal{S} \gets \text{sort}(\mathcal{S}, \text{by } s(q,p_i)\;\text{descending})$

\BlankLine
\textbf{Step 3: Evidence Filtering} \\
$\mathcal{P}^* \gets \{ p_i \mid (p_i,s(q,p_i)) \in \mathcal{S},\; s(q,p_i) \geq \tau \}$ \\
\tcp{Remove noisy or tangential passages}

\BlankLine
\textbf{Step 4: Evidence Summarization} \\
$\mathcal{E} \gets \emptyset$ \\
\For{$p_i \in \mathcal{P}^*$}{
    $e_i \gets \text{LLM}_{\text{summ}}(p_i, I(q))$ \\
    \tcp{Condense passage: keep key entities, preserve terminology}
    $\mathcal{E} \gets \mathcal{E} \cup \{ e_i \}$
}

\BlankLine
\Return $\mathcal{E}$
\end{algorithm}
 
\section{Experiment}
\subsection{Experimental Settings}

\paragraph{Dataset}
We evaluate our approach on two datasets constructed from a corpus of approximately 10M scientific articles, yielding 300K QA pairs. Dataset I is generated using the LMQG framework, while Dataset II is built with our proposed dataset construction method (Section~\ref{sec:Methodology}). 
For RAG evaluation, each dataset is split into \textit{Base} and \textit{SSLI} subsets. The \textit{Base} subset contains 75K QA instances paired with directly retrieved contexts, while the \textit{SSLI} subset includes 75K instances with semantically similar but logically irrelevant distractor passages injected into the retrieved contexts. For all experiments, we retrieve the top-30 passages per query to ensure fair comparison.

\paragraph{Baselines}
For baseline settings, we evaluate a diverse set of reranking models that represent the main paradigms in the field. 
(1) \textbf{Dense cross-encoder} models such as BGE~\cite{chen2024bge}, Jina~\cite{jina}, BCE~\cite{youdao_bcembedding_2023} are included as strong neural baselines, 
while \textbf{sparse lexical} approaches are represented by SPLADE~\cite{formal2021splade}, which provides interpretable, term-level signals. 
(2) To capture token-level \textbf{late interaction} mechanisms, we incorporate ColBERT~\cite{santhanam2021colbertv2}, and for 
(3) \textbf{sequence-to-sequence} generation, we test T5-based rerankers and MonoT5~\cite{nogueira-etal-2020-document} along with 
(4) \textbf{listwise} extensions such as RankT5~\cite{zhuang2023rankt5} and ListT5~\cite{yoon2024listt5}. 
(5) We further include \textbf{distillation-based} rerankers, such as Twolar~\cite{baldelli2024twolar}, as well as 
(6) instruction-following 
\textbf{LLM-based} rerankers including RankGPT~\cite{sun2023chatgpt}, VicunaReranker~\cite{pradeep2023rankvicuna}, ZephyrReranker~\cite{pradeep2023rankzephyr}, DeARank~\cite{abdallah-etal-2025-dear}, limrank~\cite{song2025limrank}, and LLM2Vec~\cite{behnamghader2024llm2vec}, which highlight the capability of LLMs to directly generate ranking signals. 

\paragraph{Experimental Settings.}
Our agentic reranker follows a two-stage pipeline.
An LLM first performs intention recognition to derive a structured representation of the input question, which conditions the subsequent relevance scoring of candidate passages.
Each passage is assigned a RelevanceScore, and passages with scores below a threshold ($\tau=0.8$) are filtered.
The retained passages are further summarized into concise, query-conditioned representations to reduce redundancy and context length.
For fair comparison, all methods retrieve the top-30 passages, with $k=5$ passages selected for generation.
We implement our reranker using Qwen-plus~\cite{qwen-plus}, and conduct all generation experiments with the same generator (\texttt{Llama-2-70b-chat-hf}).

\paragraph{Implementation Details.} 
All reranking experiments are performed on a cluster equipped with four NVIDIA H100 GPUs. Both LLMs and reranking models are evaluated using their default parameter settings. To reduce randomness, each experiment is repeated three times with independent runs. For consistency and fairness, all methods are assessed in a zero-shot regime, without any task-specific fine-tuning or auxiliary optimization.

\paragraph{Evaluation Metrics}
We conduct experiments using a combination of generation-based and rerank-based metrics to comprehensively assess their performance. 
For the \textbf{generation quality of the LLM answer}s, we report standard metrics including F1, Precision and Recall, which measure the overlap and completeness of the generated answer relative to the ground truth. 
For \textbf{reranking performance}, we adopt two tailored metrics: \textit{HitRate@K}, which measures whether the golden context appears within the top-$K$ retrieved passages, and \textit{Relative Position (RP)}, which captures the rank of the golden context, reflecting the model’s ability to promote relevant evidence.
However, lexical overlap metrics are insufficient under the \textit{SSLI} setting, where answers may exhibit high surface similarity yet lack correct logical grounding. To address this limitation, we introduce the \textbf{Logic Fidelity Score (LFS)}, an LLM-based metric that evaluates answers based on factual correctness, logical consistency, and contextual adequacy. Given a question $q$, golden answer $a^{*}$, and candidate answer $a$, the LLM assigns a score from $0$ to $5$, with higher values indicating stronger logical alignment with the ground truth.

\begin{table*}[htbp]
\centering
\scriptsize
\caption{Evaluation results of different reranking models on scientific QA dataset with \textit{SSLI} setting. 
Each model is evaluated under two difficulty settings: easy and hard. 
Generation metrics (left) evaluate the quality of LLM answers, while reranking metrics (right) measure retrieval effectiveness. }
\resizebox{\textwidth}{!}{%
\begin{tabular}{c|c|cccc|ccc}
\toprule
\multirow{2}{*}{\textbf{Model}} & \multirow{2}{*}{\textbf{Setting}} 
 & \multicolumn{4}{c|}{\textbf{Generation Metrics}} 
 & \multicolumn{3}{c}{\textbf{Reranking Metrics}} \\
 &  & F1 $\uparrow$ & Precision $\uparrow$ & Recall $\uparrow$ & LFS $\uparrow$ 
 & HitRate@1 $\uparrow$ & HitRate@3 $\uparrow$ & RP $\uparrow$ \\
\midrule
\multirow{2}{*}{BCE} 
    & LMQG & 46.49 $\pm$ 1.12 & 34.58 $\pm$ 0.97 & 63.82 $\pm$ 1.05 & 4.01 $\pm$ 0.88 
    & 53.80 $\pm$ 1.20 & 64.80 $\pm$ 1.02 & 59.73 $\pm$ 1.15 \\
    & Ours & \textbf{46.86 $\pm$ 1.08} & 39.33 $\pm$ 1.21 & 53.05 $\pm$ 0.95 & 3.31 $\pm$ 0.67 
    & 51.40 $\pm$ 1.18 & 73.20 $\pm$ 1.10 & 62.91 $\pm$ 1.25 \\
\midrule
\multirow{2}{*}{BGE} 
     & LMQG & 45.85 $\pm$ 0.98 & 35.53 $\pm$ 1.03 & 62.49 $\pm$ 0.92 & 4.03 $\pm$ 0.81 & 53.60 $\pm$ 1.10 & 65.20 $\pm$ 1.12 & 59.53 $\pm$ 1.07 \\
     & Ours & 46.40 $\pm$ 1.05 & 40.98 $\pm$ 1.18 & 53.40 $\pm$ 1.02 & 3.40 $\pm$ 0.78 
     & 59.40 $\pm$ 1.25 & 82.60 $\pm$ 1.10 & 71.61 $\pm$ 1.18 \\
\midrule
\multirow{2}{*}{Jina} 
  & LMQG & 46.02 $\pm$ 1.15 & \textbf{36.44 $\pm$ 0.95} & 61.36 $\pm$ 0.88 & 3.99 $\pm$ 0.73 & 52.80 $\pm$ 1.03 & 64.20 $\pm$ 1.08 & 59.04 $\pm$ 1.09 \\
  & Ours & 44.93 $\pm$ 1.10 & \textbf{43.61 $\pm$ 0.92} & 48.39 $\pm$ 1.05 & 3.61 $\pm$ 0.81 & 60.80 $\pm$ 1.14 & 76.20 $\pm$ 1.05 & 68.36 $\pm$ 1.12 \\
\midrule
\multirow{2}{*}{MiniLM} 
  & LMQG & 44.88 $\pm$ 0.97 & 33.65 $\pm$ 0.87 & 61.15 $\pm$ 1.03 & 3.99 $\pm$ 0.91 & 52.60 $\pm$ 1.07 & 63.60 $\pm$ 1.12 & 58.38 $\pm$ 1.15 \\
  & Ours & 45.25 $\pm$ 1.12 & 38.91 $\pm$ 1.10 & 51.43 $\pm$ 0.97 & 3.20 $\pm$ 0.72 & 34.20 $\pm$ 1.05 & 58.60 $\pm$ 1.08 & 47.74 $\pm$ 1.09 \\
\midrule
\multirow{2}{*}{In-Ranker} 
    & LMQG & 45.40 $\pm$ 1.05 & 33.74 $\pm$ 0.98 & 61.90 $\pm$ 1.02 & 3.98 $\pm$ 0.81 & 54.40 $\pm$ 1.10 & 65.40 $\pm$ 1.07 & 60.33 $\pm$ 1.12 \\
    & Ours & 45.27 $\pm$ 1.08 & 38.75 $\pm$ 1.15 & 53.29 $\pm$ 0.99 & 3.38 $\pm$ 0.79 & 53.40 $\pm$ 1.12 & 81.40 $\pm$ 1.08 & 67.53 $\pm$ 1.10 \\
\midrule
\multirow{2}{*}{MXBAI} 
  & LMQG & 45.64 $\pm$ 0.95 & 34.46 $\pm$ 1.01 & 62.70 $\pm$ 0.89 & 4.08 $\pm$ 0.86 & 47.80 $\pm$ 1.05 & 62.20 $\pm$ 1.08 & 55.35 $\pm$ 1.03 \\
  & Ours & 45.79 $\pm$ 1.00 & 39.76 $\pm$ 1.08 & 53.50 $\pm$ 0.91 & 3.55 $\pm$ 0.84 & 62.60 $\pm$ 1.12 & \textbf{83.20 $\pm$ 1.05} & 63.21 $\pm$ 1.10 \\
\midrule
\multirow{2}{*}{ColBERT} 
  & LMQG & 45.35 $\pm$ 1.03 & 34.73 $\pm$ 0.97 & 61.95 $\pm$ 0.95 & 4.02 $\pm$ 0.82 & 51.80 $\pm$ 1.05 & 63.40 $\pm$ 1.09 & 58.15 $\pm$ 1.07 \\
  & Ours & 45.39 $\pm$ 1.08 & 39.43 $\pm$ 1.11 & 52.52 $\pm$ 0.98 & 3.32 $\pm$ 0.80 & 48.80 $\pm$ 1.10 & 75.80 $\pm$ 1.08 & 63.24 $\pm$ 1.12 \\
\midrule
\multirow{2}{*}{RankT5} 
  & LMQG & 44.81 $\pm$ 1.02 & 33.58 $\pm$ 0.97 & 60.87 $\pm$ 0.93 & 3.99 $\pm$ 0.78 & 55.40 $\pm$ 1.08 & 65.40 $\pm$ 1.05 & 60.70 $\pm$ 1.11 \\
  & Ours & 45.48 $\pm$ 1.06 & 40.14 $\pm$ 1.09 & 52.72 $\pm$ 0.95 & 3.37 $\pm$ 0.80 & 52.60 $\pm$ 1.09 & 78.40 $\pm$ 1.07 & 65.95 $\pm$ 1.10 \\
\midrule
\multirow{2}{*}{ListT5} 
  & LMQG & 19.64 $\pm$ 1.12 & 14.64 $\pm$ 0.91 & 27.94 $\pm$ 1.05 & 2.63 $\pm$ 0.72 & 0.20 $\pm$ 0.58 & 0.20 $\pm$ 0.61 & 0.25 $\pm$ 0.59 \\
  & Ours & 21.34 $\pm$ 1.05 & 23.97 $\pm$ 1.01 & 26.52 $\pm$ 0.93 & 1.81 $\pm$ 0.67 & 0.60 $\pm$ 0.62 & 1.80 $\pm$ 0.74 & 1.20 $\pm$ 0.68 \\
\midrule
\multirow{2}{*}{SPLADE} 
  & LMQG & 45.48 $\pm$ 1.07 & 35.04 $\pm$ 0.95 & 60.68 $\pm$ 0.97 & 3.93 $\pm$ 0.80 & 48.60 $\pm$ 1.08 & 62.80 $\pm$ 1.05 & 56.07 $\pm$ 1.10 \\
  & Ours & 45.07 $\pm$ 1.03 & 38.74 $\pm$ 1.08 & 51.40 $\pm$ 0.92 & 3.28 $\pm$ 0.78 & 51.40 $\pm$ 1.09 & 75.60 $\pm$ 1.08 & 64.38 $\pm$ 1.10 \\
\midrule
\multirow{2}{*}{TwoLAR} 
  & LMQG & 45.36 $\pm$ 1.02 & 34.10 $\pm$ 0.95 & 62.68 $\pm$ 0.98 & 4.04 $\pm$ 0.81 & 48.80 $\pm$ 1.07 & 64.20 $\pm$ 1.05 & 56.76 $\pm$ 1.09 \\
  & Ours & 46.01 $\pm$ 1.05 & 38.61 $\pm$ 1.09 & 53.76 $\pm$ 0.95 & 3.43 $\pm$ 0.78 & 58.80 $\pm$ 1.08 & 83.60 $\pm$ 1.06 & 71.47 $\pm$ 1.10 \\
\midrule
\multirow{2}{*}{LLM2Vec} 
  & LMQG & 44.39 $\pm$ 1.04 & 32.46 $\pm$ 0.93 & 62.04 $\pm$ 0.98 & 4.01 $\pm$ 0.82 & 20.40 $\pm$ 0.67 & 39.40 $\pm$ 0.74 & 32.86 $\pm$ 0.71 \\
  & Ours & 45.02 $\pm$ 1.07 & 38.69 $\pm$ 1.08 & 52.36 $\pm$ 0.95 & 3.34 $\pm$ 0.79 & 53.40 $\pm$ 1.06 & 76.40 $\pm$ 1.08 & 65.90 $\pm$ 1.12 \\
\midrule
\multirow{2}{*}{RankGPT} 
  & LMQG & \textbf{46.55 $\pm$ 1.08} & 36.21 $\pm$ 0.95 & 58.64 $\pm$ 0.97 & 3.82 $\pm$ 0.79 & 47.20 $\pm$ 1.05 & 62.20 $\pm$ 1.07 & 54.64 $\pm$ 1.09 \\
  & Ours & 45.92 $\pm$ 1.06 & 40.02 $\pm$ 1.10 & 52.88 $\pm$ 0.96 & 3.84 $\pm$ 0.80 & 50.02 $\pm$ 1.08 & 70.10 $\pm$ 1.06 & 57.98 $\pm$ 1.08 \\
\midrule
\multirow{2}{*}{VicunaReranker} 
  & LMQG 
  & 38.74 $\pm$ 0.92 
  & 35.56 $\pm$ 1.49 
  & 60.73 $\pm$ 2.36 
  & 4.03 $\pm$ 0.14 
  & 45.18 $\pm$ 1.77 
  & 59.24 $\pm$ 2.48 
  & 52.15 $\pm$ 1.00 \\
  & Ours 
  & 41.12 $\pm$ 0.42 
  & 41.49 $\pm$ 2.31 
  & 54.38 $\pm$ 1.75 
  & 3.65 $\pm$ 0.12 
  & 58.67 $\pm$ 3.07 
  & 74.95 $\pm$ 1.02 
  & 67.56 $\pm$ 1.84 \\
\midrule
\multirow{2}{*}{ZephyrReranker} 
  & LMQG 
  & 38.97 $\pm$ 1.84 
  & 36.73 $\pm$ 2.13 
  & 61.03 $\pm$ 4.57 
  & 4.00 $\pm$ 0.17 
  & 50.80 $\pm$ 4.10 
  & 61.24 $\pm$ 4.30 
  & 56.39 $\pm$ 4.27 \\
  & Ours 
  & 41.00 $\pm$ 0.47 
  & 40.63 $\pm$ 2.19 
  & 54.41 $\pm$ 1.30 
  & 3.59 $\pm$ 0.09 
  & 61.69 $\pm$ 4.84 
  & 72.95 $\pm$ 3.56 
  & 69.02 $\pm$ 3.44 \\
\midrule
\multirow{2}{*}{MonoT5} 
  & LMQG 
  & 38.56 $\pm$ 1.30 
  & 35.97 $\pm$ 1.76 
  & 60.43 $\pm$ 0.91 
  & 3.92 $\pm$ 0.07 
  & 51.41 $\pm$ 2.52 
  & 64.06 $\pm$ 2.80 
  & 57.66 $\pm$ 2.03 \\
  & Ours 
  & 38.95 $\pm$ 1.01 
  & 39.08 $\pm$ 3.57 
  & 51.85 $\pm$ 1.21 
  & 3.32 $\pm$ 0.10 
  & 45.58 $\pm$ 3.98 
  & 72.29 $\pm$ 2.14 
  & 59.79 $\pm$ 2.48 \\
\midrule
\multirow{2}{*}{limrank} 
  & LMQG 
  & 33.19 $\pm$ 0.96 
  & 30.94 $\pm$ 1.64 
  & 55.06 $\pm$ 1.72 
  & 3.86 $\pm$ 0.01 
  & 14.86 $\pm$ 2.80 
  & 32.93 $\pm$ 3.82 
  & 25.95 $\pm$ 3.04 \\
  & Ours 
  & 37.44 $\pm$ 0.95 
  & 36.73 $\pm$ 4.46 
  & 52.60 $\pm$ 2.37 
  & 3.46 $\pm$ 0.03 
  & 35.14 $\pm$ 3.12 
  & 57.23 $\pm$ 3.44 
  & 47.63 $\pm$ 2.49 \\
\midrule
\multirow{2}{*}{DeARrerank} 
  & LMQG 
  & 37.85 $\pm$ 1.36 
  & 35.05 $\pm$ 1.61 
  & 61.01 $\pm$ 2.29 
  & 3.96 $\pm$ 0.07 
  & 53.01 $\pm$ 3.84 
  & 63.65 $\pm$ 3.20 
  & 58.70 $\pm$ 2.92 \\
  & Ours 
  & 39.49 $\pm$ 0.96 
  & 39.35 $\pm$ 3.28 
  & 54.04 $\pm$ 0.59 
  & 3.51 $\pm$ 0.10 
  & 61.85 $\pm$ 2.88 
  & 73.13 $\pm$ 3.55 
  & 62.90 $\pm$ 2.76 \\

\midrule
\multirow{2}{*}{\textbf{DeepEra}} 
  & LMQG & 46.00 $\pm$ 1.05 & 34.15 $\pm$ 0.97 & \textbf{64.35 $\pm$ 0.99} & \textbf{4.09 $\pm$ 0.83} & \textbf{56.20 $\pm$ 1.10} & \textbf{66.60 $\pm$ 1.08} & \textbf{63.65 $\pm$ 1.09} \\
  & Ours & 43.38 $\pm$ 1.04 & 42.97 $\pm$ 1.09 & \textbf{60.54 $\pm$ 0.95} & \textbf{3.94 $\pm$ 0.80} & \textbf{66.60 $\pm$ 1.07} & 76.40 $\pm$ 1.06 & \textbf{71.96 $\pm$ 1.08} \\
  
\bottomrule
\end{tabular}}
\label{tab:model_results}
\vspace{-8mm}
\end{table*}

\begin{figure*}[htbp]
    \centering
    \includegraphics[width=\linewidth]{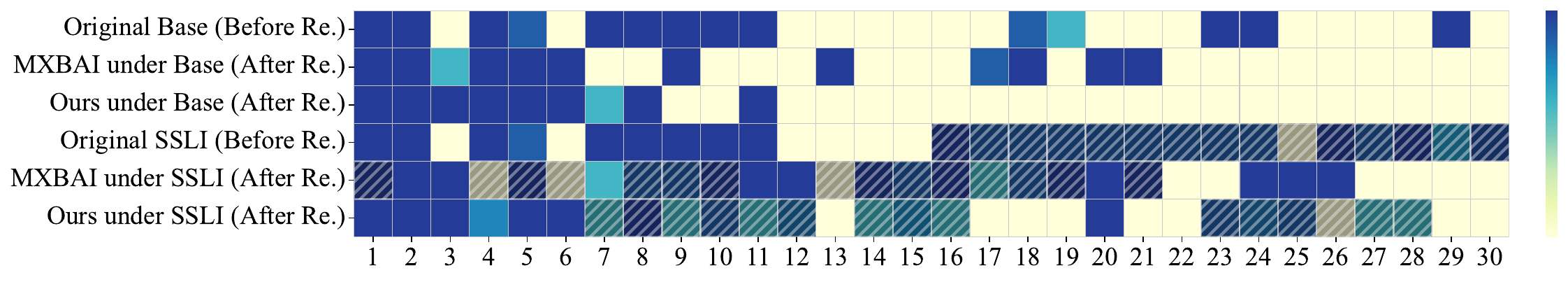}%
    \vspace{-2mm}
    \caption{Visualization of the relevance between questions and retrieved contexts across six evaluation tasks, before and after reranking. Shaded regions denote semantically similar but logically irrelevant passages.}
    \label{fig:exp_hot}
\end{figure*}

\subsection{Overall Performance}

\subsubsection{Generation Performance}

Table~\ref{tab:model_results} highlights the generation performance of various rerankers under both base and \textit{SSLI} settings. 
Precision, which measures the proportion of correctly generated content relative to all generated content, \textbf{declines significantly for most baselines}, indicating that irrelevant context can mislead the model into including incorrect information. 
Recall, reflecting the completeness of the generated answer relative to the ground truth, also varies across models; \textbf{our agentic reranker achieves the highest recall (60.54) under \textit{SSLI}}, demonstrating its effectiveness in preserving relevant evidence despite distractors. 
F1, as a harmonic mean of precision and recall, shows that our model balances accuracy and completeness better than competitors, reaching 43.38 under \textit{SSLI}.
Finally, the LLM Score, which directly evaluates answer quality and faithfulness via a separate scoring LLM, \textbf{reaches 3.94 for our model, significantly higher than other baselines}, indicating that our approach produces \textbf{more credible and trustworthy responses.} Collectively, these metrics reveal that \textbf{while \textit{SSLI} substantially increases difficulty, our method’s ability to rank relevant passages and filter out distractors} enables the LLM to generate answers that are \textbf{both accurate and comprehensive}, reflecting \textbf{robust noise resistance and contextual understanding}.

\subsubsection{Reranking Performance} 
As shown in Table~\ref{tab:model_results}, our model consistently outperforms existing baselines under both easy and hard settings. In terms of \textit{HitRate@1} and \textit{HitRate@3},\textbf{ DeepEra achieves 66.6 and 76.4 respectively on the hard subset}, surpassing the strongest baseline by over 5 points in HitRate@1 and closely matching the top performance in HitRate@3. Similarly, for the \textit{Relative Position (RP)} metric, which evaluates the ranking position of the golden context, \textbf{Ours attains 71.96 on the hard setting}, demonstrating its superior capability to \textbf{prioritize relevant passages.} These results indicate that our method is \textbf{more effective at promoting essential evidence to higher ranks}, particularly in challenging scenarios where retrieval depth is limited. 

\subsubsection{Impact of Base a \textit{SSLI} Settings}
The relative drops in \textbf{F1, Precision, and LLM Score} from Base to \textit{SSLI} quantify the difficulty introduced by semantically similar but logically irrelevant passages. Across most baselines, these reductions highlight the susceptibility of standard rerankers and generation models to noise. In contrast, our agentic reranker exhibits smaller relative declines, demonstrating its robustness to distractors. The percentage changes provide a clear picture: while \textit{SSLI} introduces additional challenging contexts, \textbf{our model effectively mitigates their negative influence}, maintaining both the fidelity and credibility of generated answers compared to other approaches.

\textbf{Recall increases for several models under \textit{SSLI}}, including our own. This can be attributed to the nature of the Recall metric, which measures the proportion of ground-truth answer tokens present in the generated response, regardless of the surrounding context. The injected distractors often contain terms semantically related to the question, which increases token overlap with the golden answer even when the overall answer quality declines. Consequently, the rise in Recall under \textbf{\textit{SSLI} should not be simply interpreted as an improvement in answer quality}; the final quality of the generated answer still needs to be assessed in conjunction with other metrics such as F1, Precision, and the LLM Score, which jointly provide a more reliable evaluation of answer correctness and relevance.

\subsubsection{LMQG vs. Ours Mode}
Evaluation under LMQG and Ours settings reveals consistent trends.
Models achieve higher performance in LMQG than in Ours, reflecting the lower difficulty of surface-level questions.
In LMQG, LLM scores often improve from Base to \textit{SSLI}, as shorter and simpler contexts enable accurate generation even with injected distractors.
Performance across rerankers also tends to converge in LMQG, suggesting limited impact of reranking for simple queries.
In contrast, Ours questions require reasoning over methods, results, and significance, leading to larger performance gaps and greater sensitivity to reranker quality.

\subsection{Visualization}
Figure~\ref{fig:exp_hot} visualizes the ranking distributions of 30 retrieved passages under the Base and \textit{SSLI} settings, where shaded regions denote semantically similar but logically irrelevant passages.
In the Base setting, both our model and MXBAI consistently rank relevant contexts at the top.
Under the more challenging \textit{SSLI} setting, MXBAI occasionally prioritizes semantically similar yet irrelevant passages, while our model maintains robust top-$5$ performance by effectively filtering misleading contexts and preserving relevant evidence.

\begin{figure}[htbp]
    \centering
    \includegraphics[width=\columnwidth]{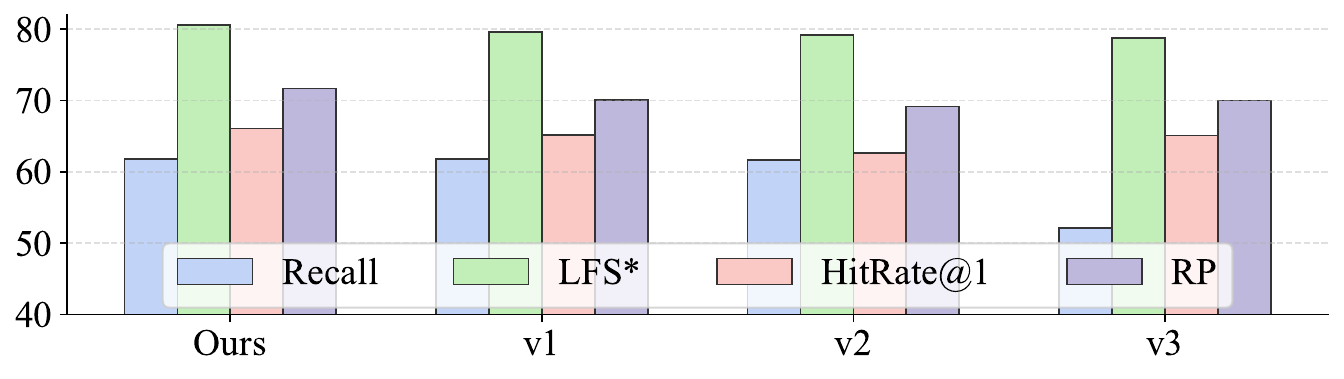}
    \vspace{-8mm}
    \caption{Ablation study of DeepEra.}
    \label{fig:ablation}
     \vspace{-2mm}
\end{figure}

\subsection{Model Analysis}
\subsubsection{Ablation Study}
To investigate the contribution of each module in our framework, we conduct ablation studies by selectively removing the three core components: (1)\textbf{v1}: removing the intention recognition module, (2) \textbf{v2}: removing the evidence filtering module, and (3) \textbf{v3}: removing the evidence summarization module. 
The results are shown in Figure \ref{fig:ablation}. V1 has a minor performance drop, with metrics decreasing slightly (e.g., HitRate@1 from 66.06 to 65.20), as the benefit of decomposing the question is limited for relatively simple queries. In contrast, v2  plays a crucial role by dynamically removing irrelevant passages; without it, additional noise is introduced, leading to noticeable declines in both generation and ranking metrics (e.g., HitRate@1 drops to 62.60). The v3  condenses each passage after reranking, so its removal has little effect on ranking indicators (HitRate@1 only drops to 65.40) but significantly reduces generation quality (Recall drops from 61.82 to 52.10, LFS from 4.03 to 3.69). Overall, these results highlight that while each module contributes differently, all three are essential for maintaining robust generation and retrieval performance in our proposed DeepEra.

\begin{figure}[t]
  \centering
  \begin{minipage}{\columnwidth}
    \centering
    \includegraphics[width=0.48\linewidth]{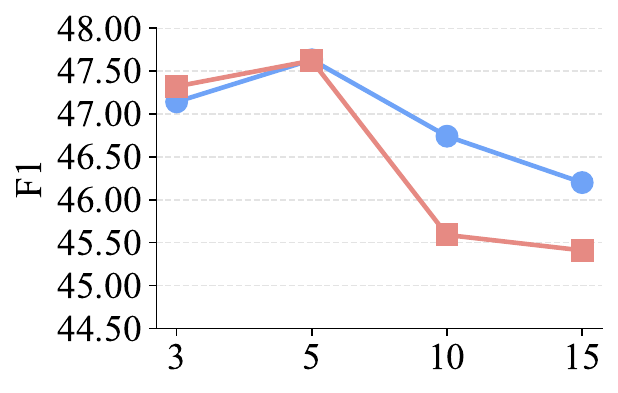}%
    \hfill
    \includegraphics[width=0.48\linewidth]{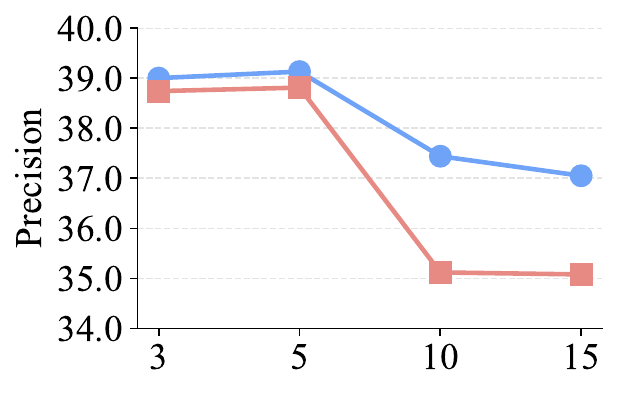}\\[-1pt]
    \includegraphics[width=0.48\linewidth]{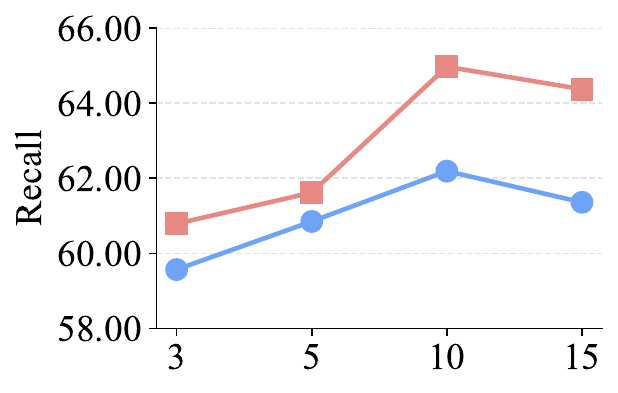}%
    \hfill
    \includegraphics[width=0.48\linewidth]{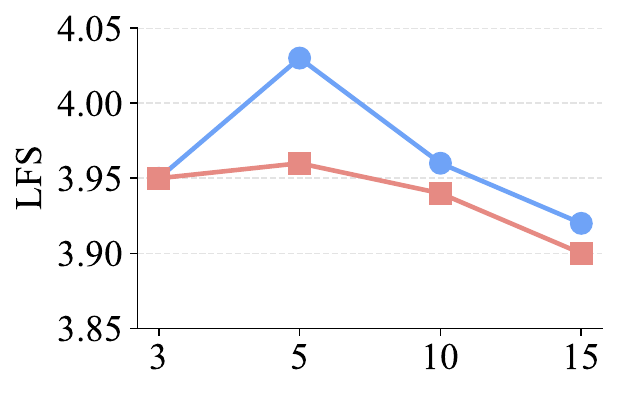}
    \vspace{-1mm}
    \caption{Parameter analysis of Top-$K$ passages in the reranking stage. Metrics are reported for DeepEra (blue) and its variant without evidence filtering (red).}
    \label{fig:parameter}
  \end{minipage}
\end{figure}

\subsubsection{Parameter Analysis}
We study the impact of varying the number of Top-$K$ passages in the reranking stage ($K = 3, 5, 10, 15$) on retrieval and generation performance, measured by HitRate@K, F1, Recall, and LFS. Fig.~\ref{fig:parameter} shows that Top-$5$ passages achieve the best balance, yielding the highest F1 and LFS while maintaining strong precision and recall. 
Second, for each Top-$K$ setting, with the \textit{Evidence Filtering} module enabled, the results are consistently higher than without filtering. This confirms the effectiveness of filtering in removing irrelevant passages, thereby improving both retrieval quality and downstream generation.  
Third, when $K$ increases beyond 10, we observe a performance drop in F1 and precision, even though recall slightly improves. This indicates that including too many passages tends to introduce redundant or noisy evidence, which negatively impacts the reranking and generation quality.

\subsubsection{Running Efficiency}
We further compare the efficiency of our framework with other LLM-based reranking methods, such as RankGPT. While RankGPT requires approximately 30 seconds to rerank a single query, our model achieves the same process in only 7.9 seconds on average. This notable improvement demonstrates that our approach not only maintains strong effectiveness but also provides a more practical solution for real-world applications where both accuracy and efficiency are crucial.

\section{Conclusion}
This paper proposes DeepEra to enhance scientific question answering by improving the selection and summarization of candidate evidence passages.
The framework integrates \textit{Intention Recognition}, \textit{Relevance Assessment}, and \textit{Evidence Summarization}.
We build two datasets to evaluate reranking robustness.
Experiments show that DeepEra effectively prioritizes relevant contexts, reduces noise, and improves answer generation quality, achieving around 8\% overall performance gain over baselines.
By filling the gap of explicit reranking in Agentic RAG frameworks, DeepEra provides a reliable, interpretable, and scalable solution for scientific QA.

\clearpage
\bibliography{cite}

\clearpage
\section{Appendix}
\subsection{Related Work}
\label{app:related_work}
\paragraph{Scientific Question Answering}
Scientific question answering (Scientific QA) aims to automatically address research or domain-specific questions by leveraging scientific texts and knowledge bases. Early approaches were largely based on information retrieval and rule-based reasoning, where candidate passages were retrieved using keyword matching and then post-processed to extract answers \cite{voorhees2001trec}. With the advent of neural architectures, sequence-to-sequence and Transformer-based readers have become dominant, enabling models to directly generate or extract answers from retrieved documents \cite{rajpurkar2016squad, devlin2019bert}.

\begin{algorithm}[htbp]
\fontsize{11pt}{13pt}\selectfont
\small
\caption{Scientific QA Dataset Construction}
\label{alg:dataset_construction}
\KwIn{Set of scientific papers $\mathcal{P}$; LLM; embedding model $\text{Embed}$; clustering function $\text{Cluster}$}
\KwOut{Dataset $\mathcal{D}$ with Base and SSLI settings}

\BlankLine
\textbf{Step 1: Abstract Segmentation and Embedding} \\
\For{$p \in \mathcal{P}$}{
    $\text{Chunks} \gets \text{segment\_abstract}(p.\text{abstract})$ \tcp*{preserve complete sentences} 
    \For{$c \in \text{Chunks}$}{
        $v_c \gets \text{Embed}(c)$ \tcp*{vector representation} 
        store\_in\_database($c$, $v_c$, $p.\text{metadata}$)
    }
}

\BlankLine
\textbf{Step 2: Corpus Clustering} \\
$\mathcal{P}^* \gets \text{Cluster}(\mathcal{P})$ \tcp*{retain contextually relevant papers}

\BlankLine
\textbf{Step 3: Structured Information Extraction} \\
\For{$p \in \mathcal{P}^*$}{
    $S_p \gets \text{LLM.extract\_structured}(p.\text{abstract})$ \tcp*{methods, results, significance}
}

\BlankLine
\textbf{Step 4: QA Pair Generation} \\
$\mathcal{QA} \gets \emptyset$ \\
\For{$S_p$ in structured info}{
    $\mathcal{QA} \gets \mathcal{QA} \cup \text{LLM.generate\_QA}(S_p)$
}

\BlankLine
\textbf{Step 5: Base and SSLI Context Construction} \\
\For{$(Q,A) \in \mathcal{QA}$}{
    $\text{BaseContexts} \gets \text{retrieve\_top\_k}(Q, \text{database}, k=3)$ \\
    $\mathcal{D} \gets \mathcal{D} \cup \{ (Q, A, \text{BaseContexts}, \text{Base}) \}$

    \tcp{SSLI: generate semantically similar but logically irrelevant distractors} 
    $\text{Guidence} \gets \text{LLM.create\_guidence}(Q, A, \text{BaseContexts})$ \tcp*{fields: DocID, TargetType, MainIdea, AnswerAvoidance} 
    $\text{Distractors} \gets \text{LLM.generate\_distractors}(\text{Guidence})$ \\
    $\text{SSLIContexts} \gets \text{BaseContexts} \cup \text{Distractors}$ \\
    $\mathcal{D} \gets \mathcal{D} \cup \{ (Q, A, \text{SSLIContexts}, \text{SSLI}) \}$
}

\BlankLine
\Return $\mathcal{D}$
\end{algorithm}

\paragraph{Two-Stage RAG Framework}
RAG integrates a retrieval module with a language generation model~\cite{yang2024crag,long2026sciencedb,zhu-etal-2025-trust}, using retrieved documents as context. Early methods employ dense retrievers (e.g., DPR \cite{karpukhin2020dpr}) or sparse retrievers (e.g., BM25 \cite{robertson2009probabilistic}). A key challenge is that retrieved passages may be noisy or irrelevant, motivating rerank models that reorder or refine results to improve their usefulness for generation \cite{nogueira2019passage}. Rerankers aim to refine the initial retrieval results by reordering candidate passages according to their relevance. 

Dense cross-encoder rerankers such as 
BGE~\cite{chen2024bge}, 
Jina~\cite{jina}, 
BCE~\cite{youdao_bcembedding_2023}, and 
MXBAI~\cite{rerank2024mxbai} 
utilize bi-sequence encoding for contextual relevance scoring, yielding strong performance. 
In contrast, sparse lexical rerankers like 
SPLADE~\cite{formal2021splade} adopts interpretable, term-level sparse representations aligned with classical IR principles. In pursuit of greater efficiency, models such as 
MiniLM~\cite{wang2020minilm} leverages distilled attention mechanisms to balance performance and cost. 
Late interaction approaches like 
ColBERT~\cite{santhanam2021colbertv2} defers token-level comparisons to maintain scalability. Recent innovations explore generative and reinforcement learning paradigms. 
LLM-based rerankers like 
RankGPT~\cite{sun2023chatgpt},
LLM2Vec~\cite{behnamghader2024llm2vec},
VicunaReranker~\cite{pradeep2023rankvicuna}, ZephyrReranker~\cite{pradeep2023rankzephyr}, DeARrerank~\cite{abdallah-etal-2025-dear},limrank~\cite{song2025limrank}
applies instruction following capabilities to generate relevance judgments, albeit at high computational cost. 
Sequence-to-sequence models like 
In-Ranker~\cite{laitz2024inranker} generates scores directly from tokenized query-document pairs, removing the need for external LLMs. 
Beyond pairwise methods, listwise rerankers such as 
RankT5~\cite{zhuang2023rankt5} and 
ListT5~\cite{yoon2024listt5} optimizes document ranking holistically through joint training. 
Two-step distillation frameworks like 
Twolar~\cite{baldelli2024twolar} efficiently transfers LLM knowledge, while 
RankR1~\cite{zhuang2025rank} introduces reinforcement learning to optimize for downstream QA accuracy directly. 

Despite progress in reranking, most models still rely on \textbf{surface-level semantic similarity}, limiting their ability to identify passages that provide genuine logical support for a query. LLM-based rerankers can partially mitigate this issue by \textbf{leveraging instruction following and in-context reasoning} to better assess relevance beyond surface similarity. However, these approaches typically \textbf{do not incorporate an explicit filtering mechanism for irrelevant passages}. As a result, semantically similar but logically misleading contexts may \textbf{still be fed into the generator}, potentially propagating errors through subsequent reasoning steps and limiting the overall reliability of the RAG pipeline. This observation motivates the need for a reranking framework that not only scores passages for relevance but also systematically filters out misleading evidence before answer generation.

\paragraph{Agentic Scientific QA}

Agentic models have recently gained attention in scientific QA,
where LLMs are equipped to plan multi-step workflows, invoke external tools, and iteratively refine answers. Representative systems such as PaperQA~\cite{lala2023paperqa} and PaSa~\cite{pasa2025} demonstrate that agentic retrieval and reasoning can substantially improve performance on scholarly QA tasks. Recent surveys further highlight the growing role of agentic AI in automating scientific discovery and literature-driven QA~\cite{survey_agent_science2025, nature_agent2025}.

Despite their success, existing agent-based QA pipelines typically \textbf{lack an explicit reranking mechanism}, allowing semantically similar but logically irrelevant passages to propagate noise into downstream reasoning and generation. To address this gap, we introduce DeepEra that systematically prioritizes logically relevant and trustworthy evidence, significantly improving the reliability and accuracy of scientific QA.

\subsection{Detailed Dataset Construction Information}
\label{sec:Appendix_Offline_Dataset}
\paragraph{Source Literature}
We construct our synthetic scientific questions using data from OpenAlex~\cite{priem2022openalex}, a large-scale open scholarly metadata repository. OpenAlex offers rich, structured metadata covering academic publications, authorship, institutional affiliations, and research concepts. The platform aggregates information from over 250 million scholarly works, includes more than 100 million distinct author entities, and catalogs metadata for approximately 120,000 publication venues spanning both journals and conferences.

\paragraph{Dataset Construction Pseudocode} Algorithm~\ref{alg:dataset_construction} presents the dataset construction workflow, encompassing abstract preprocessing, structured information extraction, QA pair generation, and the creation of Base and \textit{SSLI} settings.

\paragraph{Specific Dataset Instances/Samples}
We present representative dataset instances for both the base and SSLI settings. Each instance consists of a scientific question, its corresponding gold-standard answer, and a collection of 30 contextual passages. An example instance is illustrated in Figure~\ref{fig:dataset-sample}.

\begin{table*}[htbp]
\centering
\scriptsize
\caption{Evaluation results of different reranking models on scientific QA dataset with Base setting.  
Each model is evaluated under two difficulty settings: easy and hard. 
Generation metrics (left) evaluate the quality of LLM answers, while reranking metrics (right) measure retrieval effectiveness.}
\resizebox{\textwidth}{!}{%
\begin{tabular}{c|c|cccc|ccc}
\toprule
\multirow{2}{*}{\textbf{Model}} & \multirow{2}{*}{\textbf{Setting}} 
 & \multicolumn{4}{c|}{\textbf{Generation Metrics}} 
 & \multicolumn{3}{c}{\textbf{Reranking Metrics}} \\
 &  & F1 $\uparrow$ & Precision $\uparrow$ & Recall $\uparrow$ & LFS $\uparrow$ 
 & HitRate@1 $\uparrow$ & HitRate@3 $\uparrow$ & RP $\uparrow$ \\
\midrule
\multirow{2}{*}{BCE} 
    & LMQG & 41.37 $\pm$ 1.45 & 30.07 $\pm$ 0.88 & 60.79 $\pm$ 1.23 & 3.64 $\pm$ 0.74 & 44.51 $\pm$ 1.33 & 59.74 $\pm$ 1.20 & 52.24 $\pm$ 0.97 \\
    & Ours & 39.29 $\pm$ 0.99 & 31.67 $\pm$ 1.12 & 50.79 $\pm$ 1.41 & 2.86 $\pm$ 1.02 & 37.19 $\pm$ 1.88 & 59.77 $\pm$ 0.76 & 48.64 $\pm$ 0.89 \\
\midrule
\multirow{2}{*}{BGE} 
    & LMQG & 42.31 $\pm$ 1.05 & 31.36 $\pm$ 0.56 & 60.42 $\pm$ 1.87 & 3.62 $\pm$ 1.12 & 44.23 $\pm$ 0.77 & 59.85 $\pm$ 1.05 & 51.85 $\pm$ 0.66 \\
    & Ours & 41.60 $\pm$ 0.88 & 33.83 $\pm$ 1.11 & 51.67 $\pm$ 1.44 & 2.94 $\pm$ 0.92 & 44.72 $\pm$ 1.33 & 72.84 $\pm$ 1.02 & 58.63 $\pm$ 1.14 \\
\midrule
\multirow{2}{*}{Jina} 
    & LMQG & 44.36 $\pm$ 0.95 & 33.53 $\pm$ 1.21 & 60.14 $\pm$ 0.88 & 3.62 $\pm$ 0.77 & 44.52 $\pm$ 1.40 & 58.63 $\pm$ 1.05 & 52.77 $\pm$ 0.99 \\
    & Ours & 41.60 $\pm$ 1.23 & 38.98 $\pm$ 0.95 & 47.88 $\pm$ 1.17 & 3.44 $\pm$ 0.85 & 53.45 $\pm$ 1.22 & 71.62 $\pm$ 0.98 & 62.45 $\pm$ 1.10 \\
\midrule
\multirow{2}{*}{MiniLM} 
    & LMQG & 39.56 $\pm$ 1.11 & 28.52 $\pm$ 0.97 & \textbf{61.39 $\pm$ 1.35} & 3.63 $\pm$ 1.02 & 42.38 $\pm$ 1.09 & 57.14 $\pm$ 0.88 & 49.79 $\pm$ 0.95 \\
    & Ours & 40.57 $\pm$ 1.28 & 31.75 $\pm$ 1.12 & 49.47 $\pm$ 0.99 & 2.72 $\pm$ 1.10 & 23.58 $\pm$ 1.22 & 43.47 $\pm$ 1.17 & 34.76 $\pm$ 1.08 \\
\midrule
\multirow{2}{*}{In-Ranker} 
    & LMQG & 38.42 $\pm$ 0.88 & 29.11 $\pm$ 0.95 & 60.25 $\pm$ 1.17 & 3.68 $\pm$ 0.77 & 45.50 $\pm$ 1.11 & 58.46 $\pm$ 0.85 & 52.38 $\pm$ 1.02 \\
    & Ours & 39.60 $\pm$ 1.05 & 31.19 $\pm$ 1.02 & 50.56 $\pm$ 1.28 & 2.93 $\pm$ 0.99 & 33.80 $\pm$ 1.40 & 69.56 $\pm$ 1.18 & 51.29 $\pm$ 1.11 \\
\midrule
\multirow{2}{*}{MXBAI} 
    & LMQG & 40.11 $\pm$ 0.97 & 29.95 $\pm$ 0.88 & 61.17 $\pm$ 1.12 & \textbf{3.69 $\pm$ 0.85} & 39.40 $\pm$ 1.05 & 54.75 $\pm$ 0.88 & 47.29 $\pm$ 1.12 \\
    & Ours & 42.55 $\pm$ 1.10 & 32.79 $\pm$ 1.09 & 52.07 $\pm$ 1.28 & 3.17 $\pm$ 1.01 & 48.49 $\pm$ 1.33 & 70.61 $\pm$ 1.20 & 62.61 $\pm$ 1.09 \\
\midrule
\multirow{2}{*}{ColBERT} 
    & LMQG & 41.33 $\pm$ 1.20 & 30.43 $\pm$ 0.95 & 60.38 $\pm$ 1.33 & 3.62 $\pm$ 0.77 & 38.89 $\pm$ 0.88 & 55.11 $\pm$ 1.12 & 49.35 $\pm$ 0.97 \\
    & Ours & 40.53 $\pm$ 1.02 & 31.92 $\pm$ 0.99 & 50.79 $\pm$ 1.21 & 2.85 $\pm$ 1.05 & 28.47 $\pm$ 0.95 & 63.62 $\pm$ 1.10 & 48.91 $\pm$ 1.12 \\
\midrule
\multirow{2}{*}{RankT5} 
    & LMQG & 39.02 $\pm$ 0.88 & 28.68 $\pm$ 1.05 & 60.67 $\pm$ 1.22 & 3.61 $\pm$ 0.95 & 46.90 $\pm$ 1.10 & 61.92 $\pm$ 1.05 & 52.40 $\pm$ 1.01 \\
    & Ours & 40.61 $\pm$ 1.15 & 32.83 $\pm$ 0.95 & 51.05 $\pm$ 1.28 & 2.93 $\pm$ 0.88 & 32.88 $\pm$ 0.97 & 64.67 $\pm$ 1.08 & 51.49 $\pm$ 1.02 \\
\midrule
\multirow{2}{*}{ListT5} 
    & LMQG & 24.21 $\pm$ 0.99 & 24.01 $\pm$ 1.02 & 14.48 $\pm$ 0.77 & 1.54 $\pm$ 0.88 & 1.20 $\pm$ 0.45 & 1.20 $\pm$ 0.66 & 1.25 $\pm$ 0.51 \\
    & Ours & 20.92 $\pm$ 1.12 & 24.81 $\pm$ 0.95 & 22.34 $\pm$ 1.02 & 1.53 $\pm$ 0.78 & 0.40 $\pm$ 0.12 & 1.60 $\pm$ 0.33 & 1.00 $\pm$ 0.18 \\
\midrule
\multirow{2}{*}{SPLADE} 
    & LMQG & 42.12 $\pm$ 1.05 & 31.38 $\pm$ 0.88 & 59.84 $\pm$ 0.97 & 3.64 $\pm$ 0.77 & 36.83 $\pm$ 1.02 & 54.14 $\pm$ 1.11 & 47.33 $\pm$ 0.95 \\
    & Ours & 38.82 $\pm$ 1.18 & 30.93 $\pm$ 1.05 & 50.31 $\pm$ 1.12 & 2.69 $\pm$ 0.88 & 29.42 $\pm$ 1.12 & 63.26 $\pm$ 1.08 & 50.09 $\pm$ 1.11 \\
\midrule
\multirow{2}{*}{TwoLAR} 
    & LMQG & 40.01 $\pm$ 0.95 & 29.74 $\pm$ 0.99 & 61.36 $\pm$ 1.23 & 3.66 $\pm$ 0.77 & 37.36 $\pm$ 1.05 & 57.97 $\pm$ 0.88 & 47.89 $\pm$ 0.97 \\
    & Ours & 40.75 $\pm$ 1.02 & 31.26 $\pm$ 1.12 & 50.79 $\pm$ 1.15 & 3.01 $\pm$ 0.95 & 38.57 $\pm$ 1.17 & 74.91 $\pm$ 1.10 & 58.84 $\pm$ 1.12 \\
\midrule
\multirow{2}{*}{LLM2Vec} 
    & LMQG & 36.99 $\pm$ 1.12 & 27.67 $\pm$ 0.88 & 59.37 $\pm$ 1.11 & 3.62 $\pm$ 0.95 & 15.34 $\pm$ 0.44 & 27.22 $\pm$ 0.87 & 22.06 $\pm$ 0.66 \\
    & Ours & 38.40 $\pm$ 1.08 & 31.23 $\pm$ 1.12 & 50.73 $\pm$ 1.09 & 2.88 $\pm$ 0.88 & 39.27 $\pm$ 1.05 & 62.37 $\pm$ 1.12 & 52.31 $\pm$ 1.10 \\
\midrule
\multirow{2}{*}{RankGPT} 
    & LMQG & \textbf{45.15 $\pm$ 0.99} & \textbf{35.41 $\pm$ 1.11} & 57.12 $\pm$ 1.28 & 3.65 $\pm$ 0.88 & 47.80 $\pm$ 1.05 & 63.80 $\pm$ 0.95 & 56.08 $\pm$ 1.02 \\
    & Ours & 51.95 $\pm$ 1.05 & 51.02 $\pm$ 1.22 & 50.92 $\pm$ 1.09 & 4.00 $\pm$ 0.97 & 52.00 $\pm$ 1.08 & 74.00 $\pm$ 1.12 & 60.00 $\pm$ 0.99 \\
\midrule
\multirow{2}{*}{VicunaReranker} 
  & LMQG 
  & 42.84 $\pm$ 0.39 
  & 41.20 $\pm$ 1.92 
  & 59.18 $\pm$ 2.30 
  & 3.66 $\pm$ 0.14 
  & 47.85 $\pm$ 1.21 
  & 65.08 $\pm$ 1.18 
  & 60.14 $\pm$ 1.14 \\
  & Ours 
  & 45.27 $\pm$ 1.23 
  & 51.32 $\pm$ 1.61 
  & 52.29 $\pm$ 2.63 
  & 4.07 $\pm$ 0.12 
  & 52.53 $\pm$ 0.98 
  & 83.98 $\pm$ 1.77 
  & 58.45 $\pm$ 1.10 \\
\midrule
\multirow{2}{*}{ZephyrReranker} 
  & LMQG 
  & 41.74 $\pm$ 0.24 
  & 39.77 $\pm$ 1.33 
  & 59.29 $\pm$ 3.34 
  & 3.64 $\pm$ 0.16 
  & 43.45 $\pm$ 1.89 
  & 69.68 $\pm$ 1.01 
  & 63.64 $\pm$ 1.49 \\
  & Ours 
  & 44.22 $\pm$ 1.34 
  & 50.62 $\pm$ 2.81 
  & 51.64 $\pm$ 2.34 
  & 4.06 $\pm$ 0.11 
  & 56.35 $\pm$ 1.50 
  & 85.58 $\pm$ 0.28 
  & 60.87 $\pm$ 0.68 \\
\midrule
\multirow{2}{*}{MonoT5} 
  & LMQG 
  & 42.55 $\pm$ 0.49 
  & 40.70 $\pm$ 1.58 
  & 60.31 $\pm$ 3.53 
  & 3.63 $\pm$ 0.19 
  & 47.27 $\pm$ 1.91 
  & 60.48 $\pm$ 1.65 
  & 58.99 $\pm$ 1.62 \\
  & Ours 
  & 43.88 $\pm$ 1.06 
  & 50.63 $\pm$ 2.37 
  & 50.69 $\pm$ 1.11 
  & 3.96 $\pm$ 0.09 
  & 53.33 $\pm$ 3.69 
  & 73.78 $\pm$ 1.50 
  & 58.49 $\pm$ 2.72 \\
\midrule
\multirow{2}{*}{limrank} 
  & LMQG 
  & 37.15 $\pm$ 1.24 
  & 35.66 $\pm$ 2.53 
  & 54.15 $\pm$ 2.13 
  & 3.44 $\pm$ 0.14 
  & 40.96 $\pm$ 2.99 
  & 57.43 $\pm$ 3.28 
  & 49.74 $\pm$ 3.23 \\
  & Ours 
  & 43.12 $\pm$ 1.01 
  & 50.00 $\pm$ 2.33 
  & 49.98 $\pm$ 2.19 
  & 3.94 $\pm$ 0.03 
  & 52.05 $\pm$ 1.48 
  & 83.13 $\pm$ 0.00 
  & 62.50 $\pm$ 1.13 \\
\midrule
\multirow{2}{*}{DeARrerank} 
  & LMQG 
  & 43.47 $\pm$ 0.53 
  & 41.61 $\pm$ 1.38 
  & 60.94 $\pm$ 3.62 
  & 3.69 $\pm$ 0.18 
  & 48.07 $\pm$ 1.10 
  & 58.09 $\pm$ 1.50 
  & 50.08 $\pm$ 0.92 \\
  & Ours 
  & 44.22 $\pm$ 0.95 
  & 50.76 $\pm$ 1.73 
  & 50.67 $\pm$ 1.76 
  & 4.05 $\pm$ 0.05 
  & 46.14 $\pm$ 1.77 
  & 75.18 $\pm$ 0.85 
  & 50.61 $\pm$ 1.32 \\

\midrule
\multirow{2}{*}{\textbf{DeepEra}} 
    & LMQG & 41.11 $\pm$ 1.22 & 30.20 $\pm$ 0.97 & 60.35 $\pm$ 1.15 & 3.67 $\pm$ 0.88 & \textbf{50.00 $\pm$ 1.12} & \textbf{71.40 $\pm$ 1.08} & \textbf{66.34 $\pm$} 1.05 \\
    & Ours & \textbf{56.19 $\pm$ 1.18} & \textbf{54.86 $\pm$ 1.09} & \textbf{58.54 $\pm$ 1.02} & \textbf{4.30 $\pm$ 0.97} & \textbf{58.66 $\pm$ 1.11} & \textbf{88.63 $\pm$ 1.05} & \textbf{63.36 $\pm$ 1.12} \\
\bottomrule
\end{tabular}}
\label{tab:f1_with_deviation_base}
\end{table*}

\definecolor{lightgray}{gray}{0.95}
\begin{figure*}[htbp]
\centering
\small
\begin{tcolorbox}[colback=gray!5!white, colframe=gray!80!black, title=A  (Question-Context-Answer) Sample.]
\underline{\textbf{Question:}} How many mammal species are represented in the Yale Peabody Museum’s Division of Vertebrate Zoology collection?

\underline{\textbf{Golden Answer:}} The collection represents over 720 mammal species.

\underline{\textbf{Context (\textit{Base}):}}
\begin{itemize}
  \item \textit{Passage 1}: The mammal collection in the Yale Peabody Museums’s Division of Vertebrate Zoology, although small, is worldwide in coverage, and is used principally for teaching. The 5,086 mammal skins (over 720 species) date from the 19th century, and includes several rare and endangered species: the African elephant, black rhinoceros, orangutan, mountain gorilla, red wolf, black-footed ferret and snow leopard.
  \item \textit{Passage 2}: The Denver Museum of Nature \&amp; Science Mammal Collection currently consists of approximately 21,000 specimens, which includes 20,000+ cataloged specimens. The collection spans 1870 to the present, is worldwide in coverage, and includes specimens from the three major extant mammalian lineages (monotremes, marsupials, and placentals) distributed across 21 orders, 266 genera, and 381 species. The collection’s primary strength is its focus on the southern Rocky Mountains and Great Plains. Approximately 82\% of the collection is from the western United States, with specimens from Colorado (73\%) representing the largest percentage.
  \item \dots
  \item \textit{Passage 29}: The collection holds about 20 primary types, described as a result of California beetle surveys, housed separately from the remainder of the collection. Representation is very strong from the Transverse Ranges, including the mountains of Santa Barbara, Ventura, Kern, Los Angeles, San Bernardino, and Riverside counties. Desert areas to the east are also well represented, as is the southern Sierra Nevada.
  \item \textit{Passage 30}: Herbarium COLO is the Botany Section of the University of Colorado Museum of Natural History on the Boulder campus. The Herbarium is available to anyone with an interest in botany. Users include faculty and students, visiting scholars, private consultants, local naturalists, and botanists from a variety of public and private agencies. This dataset contains the North American portion of COLO Bryophyte collection. Approximately 50,000 specimens from the rest of the world have not been digitized. No work has been done to georeference this collection. Any coordinates provided are from the collector's label and have not been verified. 
\end{itemize}

\underline{\textbf{Context (\textit{SSLI}):}}
\begin{itemize}
  \item \textit{Passage 1}: The mammal collection in the Yale Peabody Museums’s Division of Vertebrate Zoology, although small, is worldwide in coverage, and is used principally for teaching. The 5,086 mammal skins (over 720 species) date from the 19th century, and includes several rare and endangered species: the African elephant, black rhinoceros, orangutan, mountain gorilla, red wolf, black-footed ferret and snow leopard.
  \item \textit{Passage 2}: The Denver Museum of Nature \&amp; Science Mammal Collection currently consists of approximately 21,000 specimens, which includes 20,000+ cataloged specimens. The collection spans 1870 to the present, is worldwide in coverage, and includes specimens from the three major extant mammalian lineages (monotremes, marsupials, and placentals) distributed across 21 orders, 266 genera, and 381 species. The collection’s primary strength is its focus on the southern Rocky Mountains and Great Plains. Approximately 82\% of the collection is from the western United States, with specimens from Colorado (73\%) representing the largest percentage.
  \item \dots
  \item \textit{Passage 25}: The museum's focus on historical specimens means many mammal species are represented by single specimens. This complicates accurate biodiversity assessments for the collection.
  \item \textit{Passage 26}: Yale's mammal collection includes significant holdings from Africa, but the number of species is often conflated with the larger vertebrate collection's 720 total species across all classes.
  \item \textit{Passage 27}: While the Peabody's bird collection exceeds 1,000 species, mammal diversity is less documented. Staff estimate 400-500 species based on historical acquisition records.
  \item \textit{Passage 28}: The museum's vertebrate collections emphasize geographic coverage over species count. Mammal specimens are distributed across 100+ countries but lack precise species enumeration.
  \item \textit{Passage 29}: A 2020 study cited Yale's possession of 720 endangered mammal specimens, though this refers to individual specimens rather than distinct species. Taxonomic data remains under review.
  \item \textit{Passage 30}: The Division's mammal holdings include 5,086 skins but no official species count. Cataloging efforts are currently underway to address this gap in documentation.

\end{itemize}
\end{tcolorbox}
\caption{A (Question, Context, Answer) Sample}
\label{fig:dataset-sample}
\end{figure*}

\definecolor{lightgray}{gray}{0.95}

\begin{figure*}[htbp]
\centering
\small
\begin{tcolorbox}[colback=gray!5!white, colframe=gray!80!black, title=Prompt for User Intent Recognition]
\small
\textbf{System Role:} \\
You are a scientific intent recognition agent. Your task is to analyze a scientific question and return a structured description of the user’s intent. \\

\textbf{Guidelines:}
\begin{itemize}
    \item \textbf{topic}: Identify the main scientific domain from the terminology in the question. Examples: immunology, genetics, oncology, neuroscience, metabolism.
    \item \textbf{entity\_type}: Extract the main scientific entities explicitly mentioned (cells, proteins, genes, pathways). Be specific (e.g., ``cell type: T cells''). If multiple entities are present, summarize the category (e.g., ``cell types'').
    \item \textbf{intent}: Determine the type of question based only on its phrasing. Use categories: definition, mechanism, comparison, causal, factual, functional role. Example: ``What are X and Y?'' → definition; ``How does X do Y?'' → mechanism.
    \item \textbf{expected\_answer\_type}: Infer what kind of answer the question demands. Example: ``What are the two types of lymphocytes…?'' → ``two cell types and their roles in adaptive immunity''. Example: ``How does enzyme X catalyze Y?'' → ``mechanistic explanation of catalytic process''.
\end{itemize}

\textbf{---} \\

\textbf{Human Input:} \\
\#\# Research question: \{query\} \\

\texttt{<no\_think>  <no\_think>  <no\_think>} \\

\textbf{\#\#\#} \\
Output strictly in JSON schema.
\end{tcolorbox}
\caption{Prompt used for User Intent Recognition}
\label{fig:user-intent-prompt}
\end{figure*}

\begin{figure*}[htbp]
\centering
\small
\begin{tcolorbox}[colback=gray!5!white, colframe=gray!80!black, title=Prompt for Scientific Question Generation]
\small
\textbf{System Role:} \\
You are a Question Generator Agent. Given structured notes and the original scientific passage, generate 1--3 high-quality scientific question-answer pairs (Q\&A). \\

\textbf{Guidelines:}
\begin{itemize}
    \item Questions must be self-contained, precise, and scientific. Avoid vague references like "this passage" or "the study".
    \item Cover different types: at least one method-type, one result-type, and one hypothesis/significance-type if possible.
    \item Answers must be concise but complete, directly supported by the passage and notes.
\end{itemize}

\textbf{---} \\

\textbf{Human Input:} \\
\#\#\# Structured notes from Reader Agent: \\
\{reader\_notes\} \\

\#\#\# Original scientific passage: \\
\{chunk\_text\}
\end{tcolorbox}
\caption{Prompt used for Question Generation from Structured Notes}
\label{fig:question-generator-prompt}
\end{figure*}

\subsection{Detailed Experimental Settings}

\paragraph{Detailed Prompting Template}
In this section, we present the detailed prompting templates used in our pipeline. 
Figure~\ref{fig:user-intent-prompt} presents the \textbf{User Intent Recognition} prompt, extracting a structured representation of the question. 
Finally, Figure~\ref{fig:question-generator-prompt} shows the \textbf{Scientific Question Generation} prompt, generating high-quality QA pairs from the structured Guidances and original passage.

\subsection{Detailed Experimental Results}
Detailed main results of the Base setting are shown in Table~\ref{tab:f1_with_deviation_base}. 
After correcting answer \emph{F1} with random deviations, DeepEra remains consistently superior. On the hard subset, it achieves the best scores across all generation and reranking metrics, delivering a double-digit gain in HitRate@1  over the strongest baseline and clear improvements in HitRate@3 and RP. On the easy subset, while some baselines top individual generation metrics, DeepEra still yields the best reranking effectiveness, indicating robust prioritization of golden contexts under this stricter evaluation.

\end{document}